\documentclass[letterpaper]{article} % DO NOT CHANGE THIS
\usepackage[preprint]{aaai2027}  % DO NOT CHANGE THIS
\usepackage[hyphens]{url}  % DO NOT CHANGE THIS
\usepackage{graphicx} % DO NOT CHANGE THIS
\usepackage{natbib}  % DO NOT CHANGE THIS AND DO NOT ADD ANY OPTIONS TO IT
\usepackage{caption} % DO NOT CHANGE THIS AND DO NOT ADD ANY OPTIONS TO IT
\usepackage{algorithm}
\usepackage{algorithmic}

\usepackage{newfloat}
\usepackage{listings}
\DeclareCaptionStyle{ruled}{labelfont=normalfont,labelsep=colon,strut=off} % DO NOT CHANGE THIS
\floatstyle{ruled}
\newfloat{listing}{tb}{lst}{}
\floatname{listing}{Listing}

\usepackage{booktabs}

\usepackage{xspace}

\newcommand{\method}{\textsc{VIZ-COAST}\xspace}

\title{Using VLM Reasoning to Constrain Task and Motion Planning}
\author{
    Muyang Yan\textsuperscript{\rm 1}\equalcontrib\corresponding,
    Miras Mengdibayev\textsuperscript{\rm 2}\equalcontrib,
    Ardon Floros\textsuperscript{\rm 2},
    Weihang Guo\textsuperscript{\rm 3},
    Lydia E. Kavraki\textsuperscript{\rm 3},
    Zachary Kingston\textsuperscript{\rm 2}% 
}
\affiliations{
    \textsuperscript{\rm 1} Robotics Institute, School of Computer Science, Carnegie Mellon University\\
    \textsuperscript{\rm 2} Department of Computer Science, Purdue University\\
    \textsuperscript{\rm 3} Department of Computer Science, Rice University\\
    myan2@andrew.cmu.edu, \{mmengdib, afloros, zkingston\}@purdue.edu, \{wg25, kavraki\}@rice.edu
}

\begin{document}

\maketitle


\begin{abstract}
In task and motion planning, high-level task planning is done over an abstraction of the world to enable efficient search in long-horizon robotics problems. 
However, the feasibility of these task-level plans relies on the downward refinability of the abstraction into continuous motion.
When a domain's refinability is poor, task-level plans that appear valid may ultimately fail during motion planning, requiring replanning and resulting in slower overall performance.
Prior works mitigate this by encoding refinement issues as constraints to prune infeasible task plans. 
However, these approaches only add constraints upon refinement failure, expending significant search effort on infeasible branches.
We propose \method, a method of leveraging the common-sense spatial reasoning of large pretrained Vision-Language Models to identify issues with downward refinement \emph{a priori}, bypassing the need to fix these failures during planning. 
Experiments on three challenging TAMP domains show that our approach is able to extract plausible constraints from images and domain descriptions, drastically reducing planning times and, in some cases, eliminating downward refinement failures altogether, generalizing to a diverse range of instances from the broader domain.

%In addition, we show that our method can produce constraints from few problem instances which generalize to a diverse range of instances in the broader domain.
\end{abstract}
\section{Introduction}\label{sec:intro}

Task and Motion Planning (TAMP)~\cite{garrett2021integrated} offers a framework for solving long-horizon robotic problems that couples symbolic decision-making with geometric feasibility, such as clearing a cluttered table or navigating between locked rooms using keys.
%These problems are compelling because they demand reasoning across different levels of abstraction: determining \emph{what} actions to take and \emph{how} to execute them.
% This feature makes TAMP inherently challenging as it requires reconciling discrete task structures with high-dimensional geometric constraints.
The two levels of abstraction, task and motion, enable efficient planning in long-horizon problems.
At the task level, actions are represented symbolically, such as ``pick,'' ``place,'' or ``navigate,'' with each action specified by preconditions and effects on a set of discrete predicates.
This allows the planner to reason about sequences of actions without searching in continuous geometric space.
However, these abstractions cannot express critical geometric information, violating downward refinement~\cite{bacchus1991downward,bacchus1994downward}: a valid task plan may not necessarily correspond to a feasible motion plan.
For instance, a symbolic plan may prescribe grasping an object, but that object may be obstructed, preventing any collision-free grasp.

\begin{figure}[t!]
    \centering
    \includegraphics[width=0.95\linewidth]{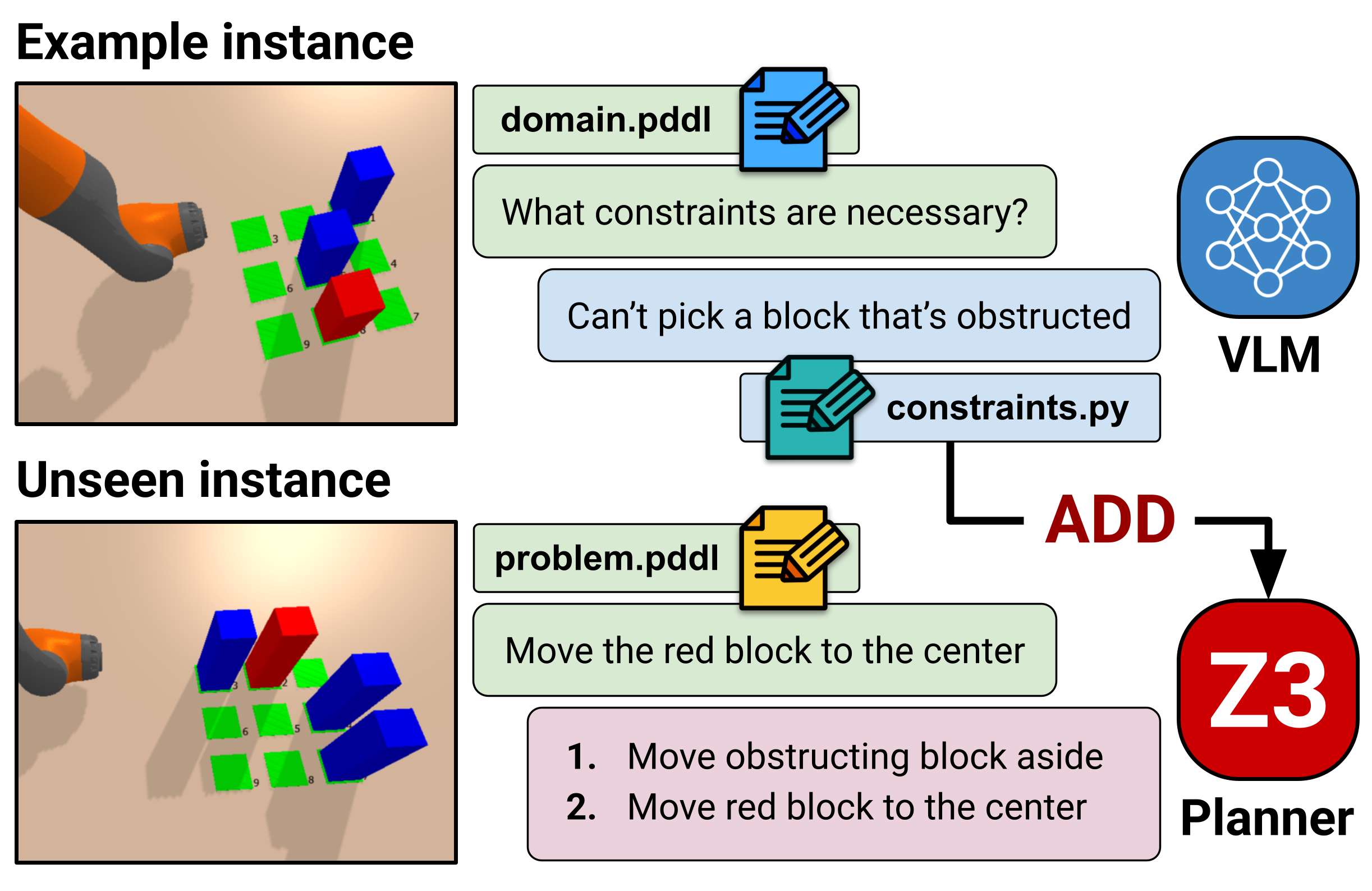}
    \caption{\method uses VLMs and SMT-based task planning to constrain TAMP problems. Our constraints generalize to unseen problem instances within a broader domain.
    \method introduces minimal search overhead, enabling more efficient planning than prior state-of-the-art.}
    \label{fig:overview}
\end{figure}

PDDLStream~\cite{garrett_pddlstream_2020} is a TAMP framework which circumvents downward refinement failures by \emph{sampling first}. Feasible continuous action parameterizations, such as grasp poses and trajectories, are sampled prior to planning via \emph{streams}, allowing task planning to be done with feasibility pre-certified.
% However, PDDLStream expends significant effort sampling groundings which may not be used.
% This can be very inefficient, as the number of possible groundings grows exponentially with object count.
However, PDDLStream expends significant effort sampling groundings that may go unused, and the number of possible groundings grows exponentially with object count.

An alternative class of methods \emph{plans first}, creating a full task plan before attempting downward refinement.
Intuitively, it is more effective to prune the search space at the discrete task level than the continuous motion level.
When refinement fails, plan-first methods translate motion failures to task-level \emph{constraints} before replanning.
For example, upon failing to find a feasible grasp pose due to an obstructing object, a constraint may be added to enforce that the offending object must not be in that position when attempting a grasp.
IDTMP~\cite{dantam2018incremental} implements this by formulating planning with SATPlan~\cite{kautz2006satplan,rintanen2014madagascar} through Z3~\cite{de2008z3}, a Satisfiability Modulo Theories (SMT) solver, incrementally adding constraint clauses upon failure.
COAST~\cite{vu2024coast}, a plan-first successor to PDDLStream, adds auxiliary predicates and entities to directly encode constraints in the PDDL description.
All these existing constraint-based approaches ultimately require expending additional search effort to encounter refinement failures and discover constraints, replanning repeatedly before finding a plan that can be ground to motion.
Our goal is to avoid expending this effort by adding constraints before planning even begins.

Recently, large pretrained vision language models (VLMs) have demonstrated common-sense geometric reasoning capabilities~\cite{gao2024physically,chen2024spatialvlm}, allowing them to generalize zero-shot to many settings.
Thus, they have been employed for various facets of TAMP, including subgoal generation~\cite{yang_guiding_2025}, preference encoding~\cite{kumar_open-world_2025}, and error correction~\cite{duan2024aha}. 

In this work, we propose \textbf{\underline{V}}isual \textbf{\underline{I}}nsight for \textbf{\underline{Z}}3-based \textbf{\underline{Co}}nstraints \textbf{\underline{a}}nd \textbf{\underline{St}}reams (\method), a novel approach to TAMP which leverages VLMs to \emph{foresee} potential downward refinement failures and add constraints to avoid them before planning begins, eliminating the need to encounter failures during planning. In challenging domains where initial constraints do not account for all failure modes, \method analyzes multiple problems to revise constraints using feedback from downstream planning in a closed-loop manner.
As shown in Figure~\ref{fig:overview}, we design a pipeline that guides a VLM to produce constraints for an SMT-based task planner. In contrast to COAST, whose auxiliary PDDL constructions introduce significant search overhead, the SMT-based solver allows us to seamlessly incorporate constraints. % while using the original PDDL description.

Through experiments on three challenging TAMP domains, we demonstrate that \method significantly decreases planning time by reducing or eliminating replanning caused by downward refinement failures. We also show that our constraints consistently generalize to unseen problem instances within a broader domain.

\section{Background and Related Work}\label{sec:related}
%We begin with a general overview of modern task and motion planning and its main challenges in~\ref{sec:tamp_bg}. We then review prior approaches in~\ref{sec:reduce_space} that accelerate TAMP by constraining the search space, highlighting the need for a \emph{generalizable} method capable of pruning \emph{before} planning. Lastly, in \ref{sec:vlm_tamp}, we explain how VLMs’ spatial reasoning capabilities make such a method possible and clarify how our approach differs from prior work.

\paragraph{Task and Motion Planning}\label{sec:tamp_bg}
Task and motion planning~(TAMP) decomposes long-horizon motion planning into search in both discrete~(task-level) and continuous~(motion-level) spaces~\cite{garrett2021integrated}. Task planning uses a symbolic abstraction of the world to produce a high-level plan from a start state to a goal condition (e.g., STRIPS~\cite{fikes1971strips}). In this work, we formalize task planning using Planning Domain Definition Language (PDDL)~\cite{mcdermott1998pddl,fox2003pddl2}.% due to its expressive power and widespread adoption.
% In task and motion planning~(TAMP), a long horizon task is task planning is done over an abstraction of the robot's capabilities, and results in sequence of actions to transition from a given start state to a desired goal condition (e.g., STRIPS~\cite{fikes1971strips}). The Planning Domain Definition Language (PDDL)~\cite{fox2003pddl2} is commonly used due to its expressive power and widespread adoption. TAMP decomposes continuous long-horizon motion planning into search in both discrete (task-level) and continuous (motion-level) spaces~\cite{garrett2021integrated}.

A core challenge in TAMP is \textit{downward-refinement failure}~\cite{bacchus1991downward,bacchus1994downward}, which occurs when a symbolic task plan becomes infeasible at the motion level.
PDDLStream~\cite{garrett_pddlstream_2020} addresses this issue by taking a \textit{sample-first} approach, performing continuous sampling before task planning via declarative streams, allowing earlier reasoning about geometric feasibility.
Other works~\cite{kaelbling_hierarchical_2011,srivastava2014combined,dantam2013motion,vu2024coast} take a \textit{plan-first} approach, designing algorithms to add \textit{constraints} upon encountering failures which inform task-level replanning to prevent revisiting infeasible options. %In the next section, we expand on the role of constraints and other methods of reducing the search space in TAMP.
% PDDLStream~\cite{garrett_pddlstream_2020} takes a different approach by moving continuous sampling from the motion level to the task level and integrating symbolic planning via declarative streams.
% This allows the planner to reason about continuous feasibility earlier and more efficiently for some problem classes.

\paragraph{Constraints in TAMP}\label{sec:reduce_space}
The purpose of \emph{constraints} in both IDTMP and COAST is to prune the task-level search space to avoid repeating failures, though they differ in implementation.
COAST~\cite{vu2024coast} is a plan-first adaptation of PDDLStream which encodes constraints by modifying the PDDL description, adding auxiliary predicates and entities to block actions under specific conditions. While this approach allows COAST to be planner-agnostic, encoding constraints directly in PDDL limits the constraints that may be expressed, and the auxiliary constructions introduce significant search overhead.
PDDL~3.0~\cite{Gerevini2005PDDLConstraints} introduced constraints on valid states, but is poorly supported by existing planners (e.g., Fast Downward~\cite{helmert2006fast}). % and cannot be reliably generated by LLMs due to limited training data.
In contrast, we follow IDTMP's~\cite{dantam2018incremental} approach and use SATPlan~\cite{kautz2006satplan,rintanen2014madagascar} implemented with the Z3~\cite{de2008z3} SMT solver, allowing constraints to be directly encoded as SMT clauses
While heuristic search approaches are typically faster~\cite{taitler20242023}, SATPlan encodings offer far greater specification flexibility and can avoid the costly translation overhead in FastDownward~\cite{helmert2006fast}.

Although existing methods~\cite{vu2024coast, dantam2018incremental} may mitigate downward refinement, they discover constraints by finding grounding failures during execution. This reactive process is repeated for many iterations before finding suitable constraints to solve a particular problem instance, and cannot generalize to new instances.

% Alternative approaches have explored \emph{predicting} action feasibility to avoid refinement failures.
% \citet{wells2019learning} propose learning feasibility constraints for tabletop manipulation tasks through domain-specific training.
% \citet{xu2022accelerating} use neural feasibility checking to identify infeasible action before motion planning.
% \citet{driess2020deep} use deep visual heuristics for learning feasibility of mixed-integer programs in manipulation planning.
% The work in~\cite{20-driess-RSS} propose using visual reasoning to predict valid actions for task and motion planning.
% While promising, these learning-based approaches require extensive domain-specific training data and may not generalize effectively to new environments or object configurations. We believe that recent advances in LLMs and VLMs enable the use of \emph{pre-trained common-sense priors} to reduce unnecessary exploration.

Alternative approaches have explored \emph{predicting} action feasibility to avoid refinement failures. \citet{wells2019learning} learn feasibility constraints for tabletop manipulation through domain-specific training, while \citet{xu2022accelerating} employ neural feasibility checks to identify infeasible actions before motion planning. \citet{driess2020deep} use deep visual heuristics to assess feasibility in mixed-integer manipulation planning, and~\citet{20-driess-RSS} leverage visual reasoning to predict valid actions for TAMP. Although promising, these methods require extensive domain-specific training data and may not generalize well to new environments or object configurations. In contrast, recent advances in LLMs and VLMs enable the use of \emph{pre-trained common-sense priors} to reduce unnecessary exploration.

% In this work, we leverage these abilities to warm-start solvers with constraints \emph{beforehand}, reducing unnecessary exploration and improving planning efficiency.

\paragraph{Vision Language Models for TAMP}\label{sec:vlm_tamp}
LLMs have demonstrated high-level reasoning~\cite{aghzal2025survey} abilities. VLMs further integrate vision and language for scene understanding~\cite{shek2024lancar}, spatial reasoning~\cite{chen2024spatialvlm}, and grounded action execution~\cite{liu2024vision, duan2024aha}. These capabilities make VLMs promising for addressing the geometric reasoning challenges in TAMP.
% Large Language Models (LLMs) have demonstrated strong capabilities in reasoning and generation, making them well-suited for high-level task planning, particularly when common sense reasoning and semantic knowledge are required~\cite{aghzal2025survey}.
% VLMs extend this capability by integrating visual and textual inputs, enabling scene understanding through recognition of objects, spatial relations, and semantic context~\cite{li2025benchmark}.
% VLM-driven methods enable translation of language instructions into grounded actions~\cite{liu2024vision} and support context-aware movement in unstructured environments~\cite{shek2024lancar}. Beyond scene understanding, recent work demonstrates VLMs' capacity for spatial and physical reasoning~\cite{chen2024spatialvlm}.
% AHA~\cite{duan2024aha} detects and explains failures in robotic manipulation to improve future planning decisions.
% These spatial reasoning capabilities make VLMs particularly promising for addressing the geometric reasoning challenges inherent in TAMP.

Several works have integrated LLMs and VLMs into TAMP pipelines for various purposes. 
\citet{zhang2023grounding} use VLMs to ground task planners by validating symbolic predicates and providing incremental feedback. \citet{hu2023look} uses a VLM to generate physically feasible actions.
ReplanVLM~\cite{mei2024replanvlm} employs VLMs for plan generation with replanning upon grounding failures.
CaStL~\cite{guo_castl_2025} translates natural language task descriptions into PDDL with constraints reflecting user preferences. OWL-TAMP~\cite{kumar_open-world_2025} encodes open-world preferences as constraints on symbolic and geometric states. Both CaStL and OWL-TAMP address problem \textit{specification}, aiming to capture preferences rather than prevent planning failures.
VLM-TAMP~\cite{yang_guiding_2025} uses VLMs to generate intermediate subgoals for long-horizon planning. While this accelerates task-level search, it presumes downward refinability.
CLIMB~\cite{byrnes_climb_2025} incrementally updates the domain description upon failure with an LLM to continually improve performance; however, it is restricted to constraints expressible in PDDL, and does not make use of visual input.

Our work leverages VLMs' spatial reasoning capabilities to proactively identify general geometric constraints that prevent downward refinement failures before planning begins.
This represents a novel application of VLM reasoning to the fundamental challenge of bridging symbolic and geometric reasoning in TAMP.
\section{Problem Formulation}\label{sec:problem}
% Our work investigates the use of VLMs to accelerate TAMP. In particular, our method produces constraints which generalize to a range of problem instances within a domain.

When inferring constraints for a particular domain, we assume access to a model of the robot and its environment, as well as a symbolic Planning Domain Definition Language (PDDL)~\cite{mcdermott1998pddl} domain description. Additionally, we assume access to some example scene within the domain, consisting of several components: (1) a symbolic PDDL problem description; (2) the geometric coordinates and dimensions of each object present; (3) an image of the initial state of the scene which clearly shows all objects. Lastly, we assume the set of possible unique object instances is finite and known.
Once constraints have been generated, they are used to plan on subsequent problem instances.

% Formally, we define a PDDL problem as a tuple $(\mathcal{P}, \mathcal{A}, \mathcal{O})$, where $\mathcal{O}$ is the finite set of objects present in the environment. These objects serve as the arguments for both predicates and actions. Each predicate function is given by $P(o_1, \dots, o_n) \in \mathcal{P}$ with $ o_i \in \mathcal{O}$ and $P: \mathcal{O}^n \rightarrow \{0, 1\}$, representing whether a relation holds for a set of objects. The action schema is defined as $A(o_1, \dots, o_n) \in \mathcal{A}$, where the action is grounded by its object arguments. 

%We use the PDDL standard for symbolic task planning. 
For a given domain, the PDDL domain description specifies the allowed types, predicates and actions.
Actions are defined by their preconditions, expressed as logical formulas over predicates, and their effects, which specify how the truth values of a subset of predicates are updated in the next state.
An instance of a problem in a particular domain defines a set of objects, their initial state, and the goal conditions. States are represented as the set of predicates which are true. We adopt the closed-world assumption, in which predicates not specified to be true are implicitly assumed to be false. The problem of task planning is to find a sequence of actions which achieves the goal from the initial state.

In TAMP, symbolic actions must also be grounded through motion planning. For example, a symbolic action such as \texttt{pick(block1, gripper)} requires the motion planner to generate a feasible trajectory and determine a valid grasp pose that enables the manipulator to successfully pick up the block. 
%This gap between task-level validity and motion-level feasibility is the core challenge posed by TAMP that we seek to address in this work, by foreseeing and imposing constraints on task-planning to improve downward-refinability.
This gap between task-level validity and motion-level feasibility is the core challenge we address.
\section{Methodology}\label{sec:method}
\vspace{1em}
\begin{figure}[!t]
    \centering
    \includegraphics[width=0.95\linewidth]{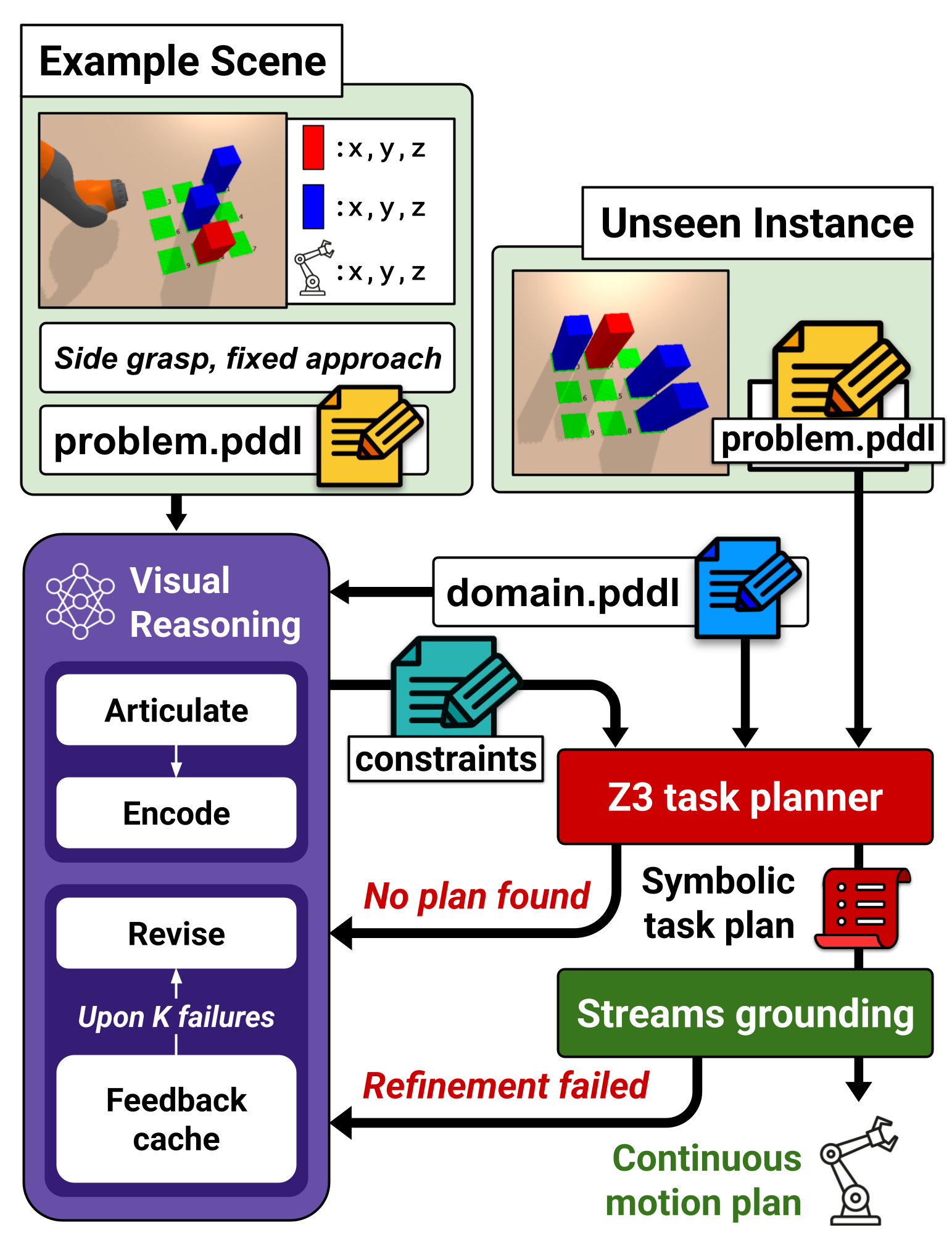}
    \caption{The \method architecture. Our Visual Reasoning Module takes as input an example scene (image and geometric state), the PDDL domain and problem description, and grasp information in natural language. It produces a Python file encoding constraints to the SMT-based task planner. Upon a new problem, the task planner takes these constraints as input, in addition to the domain and new problem description. The planner applies the constraints to block infeasible task plans. The plan produced is then grounded by a streams-based planner to produce executable motion.}
    \label{fig:architecture}
\end{figure}
% \begin{figure}
% \vspace{1em}
%     \centering
%     \includegraphics[width=\linewidth]{figures/prompting.png}
%     \caption{\method's Visual Reasoning Module infers constraints through a 4-step prompting procedure. First, it is asked to provide an interpretation of the scene based on an image of the example problem instance and the PDDL domain description. Then, it is asked to articulate in natural language the necessary constraints. After that, the VLM is asked to formally encode the constraints it identified in Python using Z3's API, referencing a structural example. Finally, it proofreads the script to ensure syntactic compliance.}
%     \label{fig:prompting}
% \end{figure}

\begin{figure}[t]
\vspace{1em}
    \centering
    \includegraphics[width=0.95\linewidth]{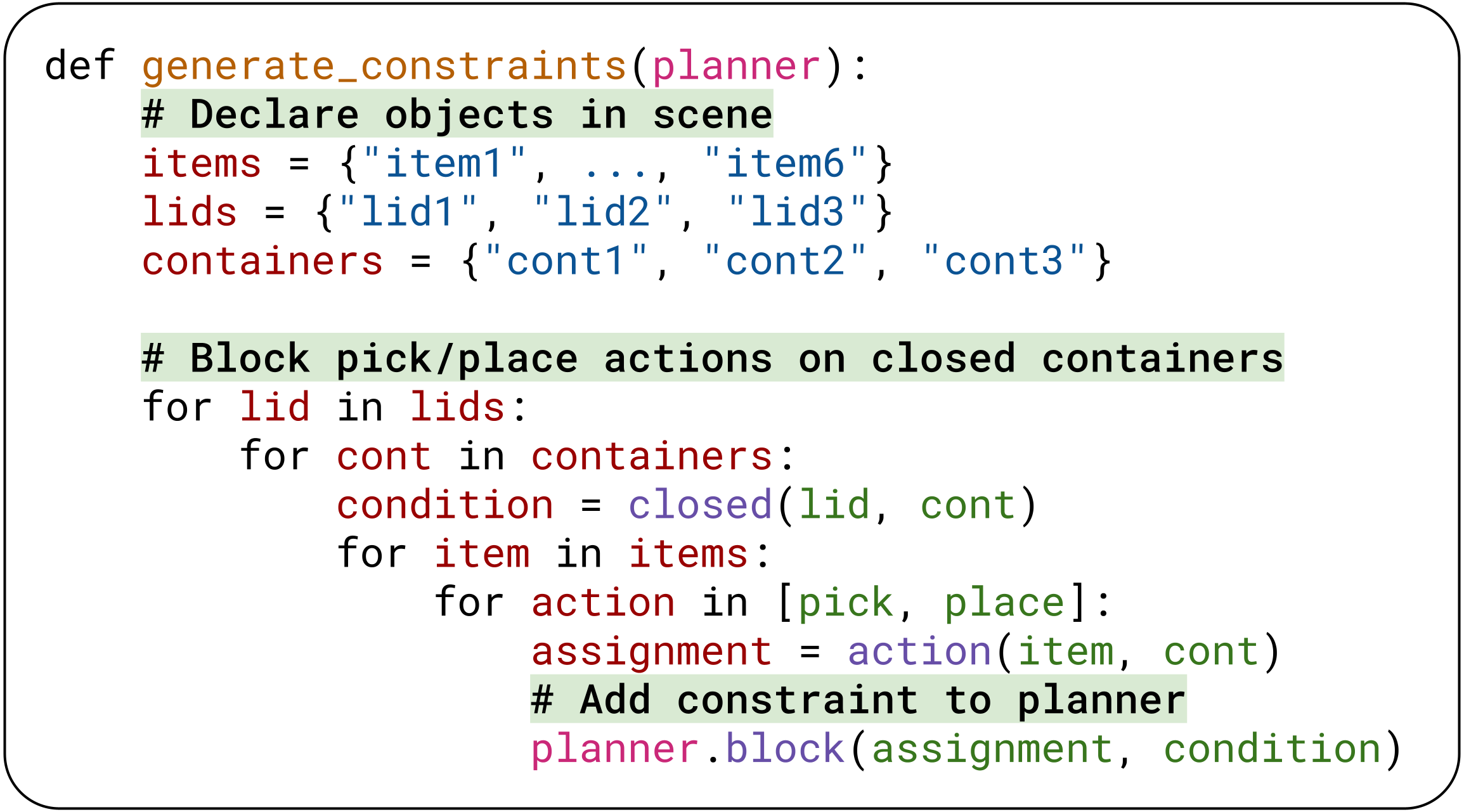}
    \caption{Paraphrased example of a constraints file produced by the VRM to block picking/placing from closed containers. The VRM produces a function which calls a high-level API to block action assignments under certain conditions, constraining the search space of the task planner.}
    \label{fig:constraints_example}
\end{figure}

\method consists of 3 primary components: 
%The task planning module, the visual reasoning module (VRM), and the motion grounding module.
\begin{itemize}
    \item \textbf{SMT-based Task planner}: We use a Satisfiability Modulo Theories (SMT) solver---Z3~\cite{de2008z3}, in particular---that uses a SATPlan or Madagascar encoding~\cite{kautz2006satplan,rintanen2014madagascar} to enable flexible constraint specification while introducing minimal search overhead. This is similar to the task planning framework used in IDTMP~\cite{dantam2018incremental} and CaStL~\cite{guo_castl_2025}.
    \item \textbf{Visual Reasoning Module (VRM)}: Before planning begins, the VRM uses a VLM to analyze an example scene to foresee downward refinement failures and infer general constraints to prevent them, encoded as constraints on the SMT solver.
    \item \textbf{Stream-based Grounding}: We ground task plans to motion with the streams framework used in PDDLStream~\cite{garrett_pddlstream_2020} and COAST~\cite{vu2024coast}.
\end{itemize}

We first briefly introduce the task planning module, then we describe the VRM in detail, and how it interacts with the downstream task and motion planning components. The full architecture is shown in Figure~\ref{fig:architecture}. Our primary contribution lies in the VRM's ability to foresee and prevent downward refinement failures. 

\subsubsection{Task planning}
\method employs a general-purpose task planner based on the Z3 SMT solver~\cite{de2008z3} using the Madagascar~\cite{rintanen2014madagascar} encoding.
This allows us to define declarative SMT constraints on allowable actions and predicates through Z3's Python API, taking the form of arbitrary expressions of first-order logic. Other planners such as Fast Downward~\cite{helmert2006fast} do not have a flexible interface for constraint specification, and require editing the PDDL files directly, introducing additional complexity especially due to the translation step to SAS+~\cite{jonsson1998state} required.
% \MuyangHighlight{We use the Madagascar~\cite{rintanen2014madagascar} encoding when possible and fall back to the SATPlan~\cite{kautz2006satplan} encoding when a domains' PDDL requirements are not supported by Madagascar.}

\subsubsection{Visual reasoning with VLMs}
\method's visual reasoning module (VRM) leverages the common-sense visual reasoning capabilities of VLMs to foresee potential refinement failures and infer necessary planning constraints. This module takes five inputs, which condition each step of prompting: PDDL specification of the problem and domain, an image of the initial state of the scene, a text-based geometric representation of the scene, a grasp specification, and an example of a compliant output. The grasp specification is a short natural-language description of how grasps and approach poses are sampled in the downstream streams-based grounding. Its output is a Python program which encodes the necessary task-level planning constraints.

The VRM consists of an initial 2-step prompting procedure, followed by a feedback mechanism. For initial constraint generation, the VLM is first asked to identify potential downward refinement issues and articulate the necessary constraints to mitigate them in \textit{natural language.} Then, the VLM is asked to encode these constraints as a Python program, using Z3's API. A paraphrased example of a constraint script generated by the VRM is shown in~\ref{fig:constraints_example}. In the event that planning still fails with these initial constraints, the VRM continues prompting the VLM in a multi-turn conversation to update the constraints using feedback from planning. There are three failure cases from planning: if the task planner fails to find a plan, the VLM is asked to make the constraints less restrictive; if the constraints raise errors during application, the VLM is given the error message and asked to debug the constraints; if a task plan is produced but results in downward refinement failure, the task plan and the stream that failed to be successfully sampled are written to a feedback cache, a temporary state-action block is added to the constraints to prevent revisiting the failed action, and planning is reattempted. If no plan succeeds after $K$ such attempts, the temporary blocks are removed and the VLM is asked to analyze the $K$ feedback entries and revise the constraints to account for these failures. This allows the VLM to reason over several unique failures of the current set of constraints to make a more informed correction.

\subsubsection{Motion Planning}
\method integrates with the COAST framework to perform downward refinement, which itself builds upon PDDLStream~\cite{garrett_pddlstream_2020}'s use of streams for sampling continuous action parameters. As the purpose of our approach is to impose task-level constraints to prevent motion planning failures, we only make modifications at the task planning level. For motion planning, we adopt COAST's stream-based approach directly without modification.

% \method's goal is to infer the necessary constraints zero-shot, before any planning failures occur. However, in the event that downward refinement still fails, we simply re-run the prompting procedure.

\section{Experiments}\label{sec:experiment} 

\subsection{Domains}
We evaluated \method on three TAMP domains, each involving robot arms manipulating objects in a simulated PyBullet environment~\cite{pybullet}:
\begin{itemize}
    \item \textbf{Shelf:} Rearrange items in a shelf such that the red block is in the center.
    \item \textbf{Packing:} Pack several items of varying shape into a container such that the lid may be placed on top.
    \item \textbf{Cooking:} Coordinate 2 arms to fry and steam ingredients and serve them on plates.
\end{itemize}

\begin{figure}[t]
\vspace{1em}
    \centering
    \includegraphics[width=\linewidth]{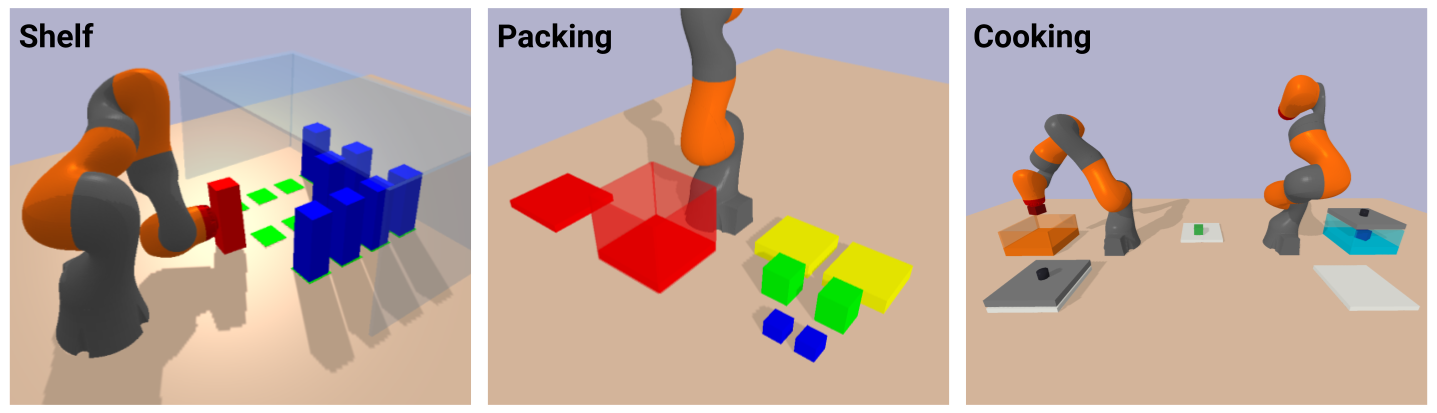}
    \caption{
    The \textbf{Shelf} domain (left) requires the robot to rearrange items in a restrictive shelf to move the red block to the center tile.
    % The \textbf{Containers} domain (middle) requires the robot to place items in target containers, removing and replacing lids as needed. 
    The \textbf{Packing} domain (middle) requires the robot to cleverly pack all items into a container and close the lid.
    The \textbf{Cooking} domain (right) requires the bimanual arms to use frying and steaming pots to cook and serve ingredients.
    }
    \label{fig:domains}
\end{figure}

Each domain has a corresponding PDDL description for task planning.
%This symbolic description cannot fully represent key considerations for downward refinement, which manifest as motion grounding failures.
For each domain, we describe the failure modes and demonstrate that \method infers constraints that not only mitigate these failures more efficiently than prior approaches, but also generalize across the class of problem instances posed by the domain.

\subsection{Baselines}
\method enables faster planning than prior state-of-the-art through reduced replanning, as well as higher success rates.
We demonstrate this by evaluating \method against several baselines and ablations:
\begin{itemize}
    \item \textbf{COAST+FD:} The original COAST system, which uses the Fast Downward task planner.
    \item \textbf{COAST+Z3:} Our novel adaption of COAST to interface with the SMT-based planner used in \method.
    \item \textbf{Incremental:} A naive learning-from-failure method which adds a single state-action block to the SMT-based planner upon refinement failure, and iteratively extends task planning horizon, similar to IDTMP~\cite{dantam2018incremental}.
    \item \textbf{VLM-only:} A baseline which prompts the VLM to directly output the task plan, given the PDDL domain and problem and an image of the initial state. Upon failure, it re-prompts with VIZ-COAST's feedback content added to the context.
    \item \textbf{No feedback:} An ablation of \method which does not use the feedback mechanism, instead starting a new conversation from scratch upon refinement failure. 
\end{itemize}
Both COAST+FD and COAST+Z3 employ COAST's approach for adding constraints upon encountering downward refinement failures.
We used Gemini 3.5 Flash with medium reasoning for all VLM calls in Shelf and Packing, and Gemini 3.6 Flash with high reasoning for Cooking. These models were selected arbitrarily based on our estimated difficulty of the domain. A comparison of different VLM models on Shelf is presented in Tbl.~\ref{tab:vlm-comparison}.

\begin{figure}
\vspace{1em}
    \centering
    \includegraphics[width=1\linewidth]{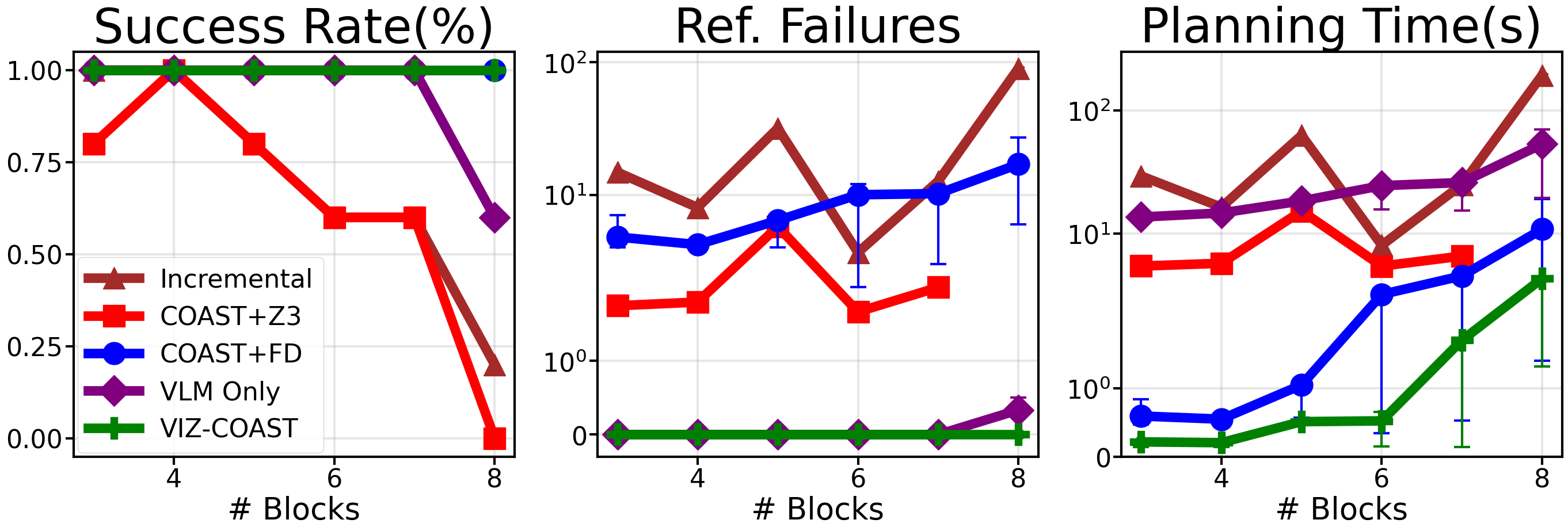}
    \caption{Success rates, refinement failures, and planning times for the \textbf{Shelf} domain. IQR bars are shown for the 3 most successful methods. \method was prompted once on the first 3-block problem, producing constraints which generalized to all 30 unique problems containing up to 8 blocks. \method plans faster than COAST-based baselines while avoiding refinement failures.}
    \label{fig:shelf_results}
\end{figure}

\subsection{Evaluation}
\textbf{Success rate} is the proportion of problems for which the method finds a successful task-and-motion plan.
\textbf{Ref. failures} refers to the total number of times each method replanned due to a downward refinement failure. Each refinement failure incurs the cost of attempting task planning and motion grounding again.
\textbf{Planning time} is the cumulative time spent task planning to solve a particular problem. It does not include motion grounding time, as the methods evaluated have little effect on the time taken to ground each task plan. It also does not include VRM prompting time, due to its variability by model. Table~\ref{tab:vlm-comparison} provides a study of the overhead of VRM calls. 

If, for a particular problem, a method exceeded 100 refinement failures or a total timeout of 600 seconds, the instance was deemed a failure. 
Metrics at each problem size for each method are averaged over all problems for which that method succeeded.
The re-prompt threshold $K$ was selected for each domain from candidates $\{1,2,5\}$ based on which yielded the fewest total refinement failures.
VLM system prompts and source code are included in Appendices.
All experiments were run on a VM with 12 vCPUs and 62 GB of RAM, hosted on an AMD Ryzen Threadripper PRO 5965WX (3.8 GHz base), with Ubuntu 24.04 LTS (Linux 7.0.0) and CPython 3.10.19.

\paragraph{Shelf}

The Shelf domain consists of a $4\times3$ grid of tiles on which blocks are placed inside a shelf facing the robot.
In each problem instance, the number of blocks and their initial configuration varies, but the goal is always to place the single red block on the tile in the middle of the second row from the robot.
The shelf restricts the robot to approach and grasp from a single direction, as shown in Figure \ref{fig:domains}.
% This information is given to the VRM as part of the grasp specification.

Because the robot's approach direction is restricted, any block obstructing the target along the direction of approach will result in a collision.
Therefore, constraints must be generated to disallow pick and place actions if an approach corridor is not clear; this is not encoded within the domain and can only be recovered by reasoning or refinement failure.

Results for the Shelf domain are shown in Figure \ref{fig:shelf_results}.
Each method was evaluated on problems with 3-8 blocks, with 5 problems per number of blocks, for a total of 30 unique problems.
Evaluation proceeded sequentially from 3-block to 8-block instances.
%To ensure a fair comparison, only data points for which at least 3 methods succeeded were included in the average.
\method's VRM constraint generation pipeline is called a single time on an example scene extracted from the first 3-block problem instance.
Constraints persist over trials until updated through the feedback mechanism.
% until $K$ motion grounding failures occur, at which point the VRM's failure feedback mechanism updates the constraints.
% In this domain we used re-prompt threshold $K=1$, such that any failure would immediately trigger a constraint update.

%comment on results
Our experiments show that on the Shelf domain, \method produces successful constraints on the initial call, before any planning occurs, which generalize to all subsequent trials over all block counts, encountering zero downward refinement failures (The No feedback ablation is thus equivalent).
In contrast, COAST-based baselines often encountered as many as 10 failures before finding a successful plan on many-block instances, resulting in slower planning.

\paragraph{Packing}

\begin{figure}
    \centering
    \includegraphics[width=1\linewidth]{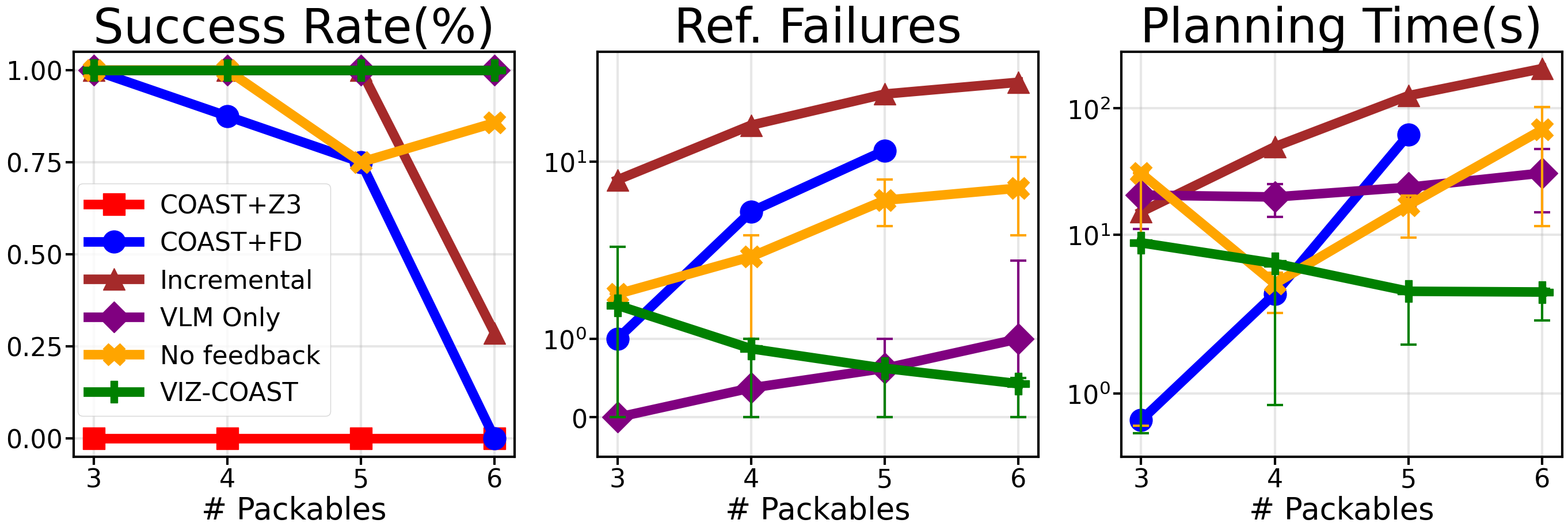}
    \caption{Success rates, refinement failures, and planning times for the \textbf{Packing} domain. IQR bars are shown for the 3 most successful methods. \method achieves 100\% success and fastest planning times on larger problem instances, whereas COAST-based methods become increasingly slower or fail completely. The failures of \method's no-feedback ablation on larger problems shows the importance of closing the constraint generation-planning loop in the VRM. As \method refines its constraints over problems, it encounters less refinement failures and plans faster.}
    \label{fig:packing_results}
\end{figure}
%Objects + overview
The Packing domain consists of 1 container with a lid and 3 kinds of packable items: slabs, big blocks, and small blocks.
In each problem, all items start on the table, and the goal is to pack them all into the container and then place the lid on top.
%what is the constraint to infer
The symbolic domain description does not account for container's limited internal volume, thus without constraints a trivial task plan is to pack all objects directly on the container floor. However, this is often geometrically infeasible as 1 slab occupies the entire floor. Constraints must be added to force stacking the items in a configuration such that they will all fit horizontally without exceeding the vertical limit, so that the lid can be placed on top.

Results for the Packing domain are shown in Figure \ref{fig:packing_results}.
Each method was evaluated on a fixed set of problem instances of 3, 4, 5, and 6 packable items where problem counts were 7, 8, 8, and 7, respectively, for a total of 30 unique instances.
\method used a re-prompt threshold of $K=5$.

%comment on results
Our results show that \method consistently has the lowest planning times on problem instances with 5 or 6 items to pack. In contrast, COAST+FD is fast on easier problems, but quickly begins to slow down and fail as complexity increases, failing completely on 6-item problems. COAST+Z3 fails on all problems. The VLM-only baseline achieves 100\% success and comparable refinement failures to \method, although it takes much longer to plan.

Unlike in Shelf, \method was unable to find fully-generalizable constraints on the first attempt, requiring continued prompting using feedback from subsequent problems to correct constraints.
The No feedback ablation failed on several 5 and 6-item problems, and encountered more refinement failures as the number of items increased. In contrast, \method's refinement failures and planning times \textit{decreased}, indicating that, through the feedback mechanism, constraints improved as more problems were attempted.

\paragraph{Cooking}

\begin{figure}
    \centering
    \includegraphics[width=1\linewidth]{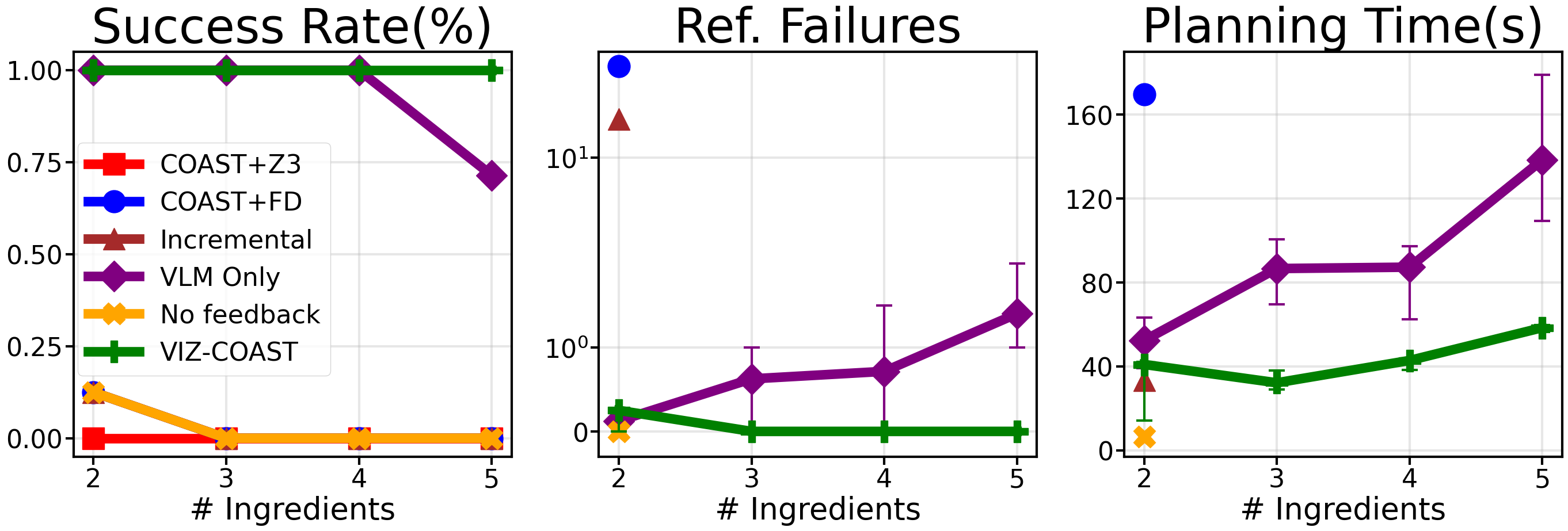}
    \caption{Success rates, refinement failures, and planning times for the \textbf{Cooking} domain. IQR bars are shown for the 2 most successful methods. \method achieves 100\% success and fastest planning times, whereas COAST-based methods and No feedback fail completely beyond 2 ingredients.
    % add a bit here
    }
    \label{fig:packing_results}
\end{figure}
%Objects + overview
The Cooking domain consists of two robot arms, one on the left and one on the right, each with a plate and pot beside them. Each pot starts covered by a lid. The left pot is for frying, and will fry an ingredient when it is placed inside. The right pot is for steaming, and will steam all ingredients inside it when a lid is placed on top. Between the two arms is a shared middle plate.
In each problem, a set of uncooked ingredients start on the left and right plates, and the goal is to transform each ingredient to its desired condition (fried, steamed or remaining uncooked) and place it on its desired plate (left, right or middle). Ingredients may not be placed on the table.
%what is the constraint to infer
This is a complex domain with several potential downward-refinement blind spots. The symbolic domain description neglects several physical properties:
(1) Ingredients cannot be picked from or placed into pots that are covered with by lids. 
(2) The left pot and plate is out of reach of the right arm, and vice versa.
(3) Lids cannot be stacked stably on top of each other due to their small protruding handles.
(4) A lid is the same size as a plate, thus ingredients and other lids cannot be placed on a plate that is occupied by a lid.
(5) Picking up a lid with ingredients on it risks the ingredients falling off.
Results are shown in Figure \ref{fig:packing_results}.
Each method was evaluated on a fixed set of problem instances of 2, 3, 4, and 5 ingredients where problem counts were 8, 8, 7, and 7, respectively, for a total of 30 unique instances.
\method used re-prompt threshold of $K=1$.

%comment on results
\method is the only method that achieves a 100\% success rate. VLM-only is able to solve most problems with higher planning times, and all other baselines fail completely beyond 2 ingredients. Notably, the No feedback ablation also exhibited this failure.
% We speculate that due to the number of downward-refinement pitfalls, the VLM is unable to foresee all potential failures from a single image of the initial state. Feedback allows it to examine the states reached during planning to uncover more subtle failures, making it essential for this domain.

\subsection{VRM Robustness}\label{sec:robustness}
\begin{table}[t]
\footnotesize
    \centering
    \resizebox{\linewidth}{!}{%
    \begin{tabular}{lrrr}
    \toprule
    VLM & Success & Prompting(s)& Ref. Failures\\
    \midrule
    Gemini3.5-Flash-High & $\textbf{100\%}$ & $49.41 \pm 8.36$ & $0.70 \pm 0.57$ \\
    Gemini3.5-Flash-Med & $\textbf{100\%}$ & $39.91 \pm 9.61$ & $0.55 \pm 0.94$ \\
    Gemini3.5-Flash-Low & $\textbf{100\%}$ & $22.87 \pm 4.37$ & ${\bf 0.10} \pm {\bf 0.31}$ \\
    Gemini3.5-Flash-Lite-Low & $\textbf{100\%}$ & ${\bf 6.28} \pm {\bf 2.20}$ & $7.05 \pm 8.87$ \\
    GPT5.6-Terra-High & $\textbf{100\%}$ & $81.17 \pm 26.70$ & $0.45 \pm 0.51$ \\
    GPT5.6-Luna-Med & $\textbf{100\%}$ & $55.31 \pm 36.85$ & $0.85 \pm 1.23$ \\
    GPT5.6-Luna-Low & $\textbf{100\%}$ & $25.24 \pm 9.77$ & $2.22 \pm 6.03$ \\
    Qwen3.7-Plus-Med & $\textbf{100\%}$ & $236.15 \pm 84.42$ & $0.35 \pm 0.49$ \\
    Qwen3.5-Flash-Med & $73\%$ & $241.83 \pm 108.52$ & $1.38 \pm 0.89$ \\
    \bottomrule
    \end{tabular}
    }
    \caption{Comparison of success rates, average time-per-prompt, and average total downward refinement failure counts of various VLM models over 20 independent trials of 4 sequential 4-7 block problems in the Shelf domain.}
    \label{tab:vlm-comparison}
\end{table}
To evaluate the consistency with which the VRM infers successful constraints, we conduct an additional comparison in which we run 20 independent trials over 4 sequential problem instances progressing from 4 to 7 blocks in the Shelf domain for various VLM configurations.
We evaluated 6 VLM models at various reasoning levels across 3 providers, with reprompt threshold $K=1$. 
We report each VLM's success rate (Success), average time per constraint update (Prompting), and average total number of refinement failures encountered over all 4 problems in a trial (Ref. Failures). Metrics are averaged over runs in which all problems succeeded.

All models achieved a 100\% success rate except for Qwen-3.5-Flash.
The Gemini family of models performed best overall, with Gemini-3.5-Flash-Low yielding the fewest refinement failures. Among GPT models, increased reasoning corresponded with a reduction in refinement failures. However, for Gemini 3.5 Flash, we see a slight opposite trend.
% We speculate that because the Shelf domain has only one mode of failure in downward refinement, increasing the reasoning length may lead the model to ``overthink," and identify extraneous constraints, indicating that \method works best when the choice of VLM matches the complexity of the domain.

\section{Discussion}

\paragraph{The Role of VLMs in TAMP}\label{sec:vlm_role}
%basically we min-max each component
The motivation behind TAMP is that efficient planning requires the right abstraction. We posit that the success of \method stems from assigning VLMs a role aligned with their strengths: synthesizing constraints to augment the symbolic abstractions used in TAMP. The disparity in performance of the VLM-only baseline across domains illustrates this point. In Packing, this baseline achieved a 100\% success rate, whereas in Shelf and Cooking its success rate dropped as problem complexity increased. We attribute this discrepancy to differences in what makes each domain challenging: in Packing, once key constraints are recognized, planning becomes straightforward, whereas Shelf and Cooking still require long-horizon, non-ergodic search, at which VLMs remain inefficient. By isolating the VLM to the constraint inference while delegating search to a symbolic planner, \method leverages the geometric intuition of VLMs while retaining the efficiency and precision of classical planning. 

More broadly, \method belongs to a growing class of methods that use VLMs to make TAMP more tractable, complementing classical planners with knowledge that is difficult to encode symbolically. Integrating constraint synthesis with complementary strategies like VLM-TAMP~\cite{yang_guiding_2025}, which uses VLMs to propose subgoals to reduce the search space for long-horizon tasks, may be future work.

\paragraph{VLMs for Constraint Synthesis}\label{sec:vlm_limits}
%summarize results, CoT good, prompting fast/practical
Table~\ref{tab:vlm-comparison} shows that VIZ-COAST performs robustly across a variety of VLM providers, models, and reasoning budgets. In general, increasing reasoning and model size yields more consistent inference of correct constraints.
% , suggesting that reasoning improves the model’s ability to extract the geometric structure required for downward refinability. 
At the same time, fast models such as Gemini-3.5-Flash still achieve strong performance while requiring as little as 20 seconds to generate constraints.
% In practice, users may choose an appropriate model and reasoning budget based on the complexity of their domain and the desired tradeoff between inference time and reliability.

%talk about how it still isn't perfect, and how feedback is necessary
\method aims first to act as a preventative method, proactively foreseeing and avoiding refinement failures.
However, as our experiments show, VLMs are not always able to foresee all failure modes. Thus, when initial constraints fail, \method uses feedback from planning. 
The Packing and Cooking domains show that this feedback is not merely helpful for recovering from occasional errors, but is often necessary for arriving at the correct insight. On Cooking, the no-feedback ablation failed completely on instances with 3 or more ingredients, whereas full \method succeeded at all problems. On Packing, the ablation saw an increase in refinement failures as it progressed through more difficult problems, whereas \method's failures decreased. These results suggest that the feedback mechanism successfully refined constraints over time by examining unforeseen failure modes encountered during planning.

% We speculate that due to the number of downward-refinement pitfalls, the VLM is unable to foresee all potential failures from a single image of the initial state. Feedback allows it to examine the states reached during planning to uncover more subtle failures, making it essential for this domain.
%this is also the motivation behind the size K feedback cache

\paragraph{The Advantage of SMT-based Constraints}
One factor in the high failure rates of COAST-based baselines in the Packing and Cooking domains is the mechanism used to encode constraints. In COAST, injecting constraints requires introducing overhead into the PDDL description, including timesteps and auxiliary logging and blocking predicates. These substantially increase the domain's symbolic complexity. In other words, COAST's mechanism for encoding constraints enables failure-driven constraint discovery \emph{at the expense of the utility of its abstraction}.
% expanding the grounded search space and introducing overhead in the translation step from PDDL to the SAS+ representation used by Fast Downward.

In contrast, \method encodes constraints directly as clauses in the intermediate SMT representation, allowing the planner to use the original PDDL description and adding minimal overhead. In the Packing and Cooking domains, this difference in representation determined whether the planner efficiently found a solution or failed entirely.

% This observation illustrates the computational advantage of SMT-based constraints while simultaneously highlighting a limitation of failure-driven constraint discovery: the process of encountering failures may itself be intractable.

% This is evidenced by the Incremental baseline, which succeeded on more problems than 
\section{Conclusion}\label{sec:conclusion}
We have presented \method, a TAMP approach that uses VLMs' common-sense geometric reasoning capabilities to constrain TAMP problems.
Unlike prior works, the constraints produced generalize to unseen problem instances and incur minimal overhead by directly interfacing with an SMT-based task planner.
These constraints prevent downward refinement failures when grounding the symbolic plan.
Furthermore, \method's closed-loop mechanism for updating constraints with planner feedback enables reasoning over several problem instances to solve difficult domains.
Our approach improves planning times and success by reducing downward refinement failures and adding constraints with minimal overhead.
For example, in the Shelf domain, \method produced constraints on the first attempt which generalized to all 30 problem instances.
In the Packing and Cooking domains, \method solves all problems, while baselines often fail to find solutions due to the complexity introduced by auxiliary constructions.
Finally, we empirically demonstrate that the performance of \method is robust to the VLM model used, and that current services are fast and powerful enough to be practical.

Overall, \method demonstrates that VLMs can provide valuable insight into problem structure to accelerate TAMP.
Future applications may plan using richer abstractions, operate in dynamic settings, or use constraints based on natural language instructions grounded to the physical world.
Although we have validated VLM models across several providers, our reliance on a proprietary cloud-based service remains a source of variance.
Evaluating \method with local or distilled models is a direction for future work.
Additionally, the generalization capacity of inferred constraints is currently limited to problems within the same PDDL domain. Future work could explore transferring or composing constraints learned from simpler domains to solve more complex ones.
Last, we evaluated our approach exclusively in simulation.
While we believe the generality of VLMs will enable constraint synthesis in many settings, empirical validation of \method's ability to handle the geometric and perceptual complexity of the real world remains future work.

\appendix

%\section*{Acknowledgments}

\bibliography{references}


\end{document}